\documentclass[chi_draft]{sigchi}

\CopyrightYear{2016}
\setcopyright{acmlicensed}
\doi{http://dx.doi.org/10.475/123_4}
\isbn{123-4567-24-567/08/06}
\conferenceinfo{CHI'16,}{May 07--12, 2016, San Jose, CA, USA}
\acmPrice{\$15.00}

\usepackage{balance}       
\usepackage{graphics}      
\usepackage[T1]{fontenc}   
\usepackage{txfonts}
\usepackage{mathptmx}
\usepackage[pdflang={en-US},pdftex]{hyperref}
\usepackage{color}
\usepackage{booktabs}
\usepackage{textcomp}

\usepackage{microtype}        
\usepackage{ccicons}          

\usepackage[textsize=small]{todonotes}
\usepackage{amsmath}
\usepackage{empheq}
\usepackage{mdframed}

\newlength\dlf  

\def\plaintitle{On Attention Models for Human Activity Recognition}

\def\emptyauthor{}
\def\plainkeywords{Authors' choice; of terms; separated; by
  semicolons; include commas, within terms only; required.}

\makeatletter
\def\url@leostyle{%
  \@ifundefined{selectfont}{
    \def\UrlFont{\sf}
  }{
    \def\UrlFont{\small\bf\ttfamily}
  }}
\makeatother
\def\pprw{8.5in}
\def\pprh{11in}

\definecolor{linkColor}{RGB}{6,125,233}
\hypersetup{%
  pdftitle={\plaintitle},
  pdfauthor={\emptyauthor},
  pdfkeywords={\plainkeywords},
  pdfdisplaydoctitle=true, 
  bookmarksnumbered,
  pdfstartview={FitH},
  colorlinks,
  citecolor=black,
  filecolor=black,
  linkcolor=black,
  urlcolor=linkColor,
  breaklinks=true,
  hypertexnames=false
}

\begin{document}

\title{\plaintitle}

\numberofauthors{2}
\author{%
  \alignauthor{Vishvak S Murahari\\  
    \affaddr{Georgia Institute of Technology}\\
    \affaddr{Atlanta, USA}\\
    \email{vishvak.murahari@gatech.edu}}\\
  \alignauthor{Thomas  Pl\"{o}tz\\
    \affaddr{Georgia Institute of Technology}\\
    \affaddr{Atlanta, USA}\\
    \email{thomas.ploetz@gatech.edu}}\\
}

\maketitle

\begin{abstract}
Most approaches that model time-series data in human activity recognition based on body-worn sensing (HAR) use a fixed size temporal context to represent different activities. 
This might, however, not be apt for sets of activities with individually varying durations. 
We introduce attention models into HAR research as a data driven approach for exploring relevant temporal context. 
Attention models learn a set of weights over input data, which we leverage to weight the temporal context being considered to model each sensor reading. 
We construct attention models for HAR by adding attention layers to a state-of-the-art deep learning HAR model (DeepConvLSTM) and evaluate our approach on benchmark datasets achieving significant increase in performance. 
Finally, we visualize the learned weights to better understand what constitutes relevant temporal context.
\end{abstract}

\category{H.1.2.}{User/Machine System}{} 
\category{I.5.}{Pattern Recognition}{}



\section{Introduction}

In Human Activity Recognition (HAR) we analyze and model sequential, that is time-series data. 
In order to do so we need to look into the temporal context of every single sensor reading, which forms the basis for modeling and eventually recognition. 
This has traditionally been done through sliding window approaches \cite{Bulling:2014:THA:2578702.2499621}, which use a fixed size window to model the temporal context of every single sensor reading.
Sliding window procedures also (and still) play a crucial role for many recent Deep Learning based HAR methods.
For example, Convolutional Neural Networks (CNNs) in HAR employ a sliding window procedure to map the timeseries data to a fixed 2D representation that is fed into the convolution layers \cite{ordonez2016deep}.
Arguably, the window length is a crucial parameter for sliding-window based approaches that often is established based on prior, i.e., domain knowledge.
Decisions regarding any sliding window procedure are hard and often final decisions that impact the recognition procedure as a whole.
As such mistakes made here are critical and errors made are difficult to recover from.
Also, most sliding-window based approaches are constrained to use a single fixed size context, which may not be ideal when modeling activities with varying durations.

Alternative approaches use sequential models that could overcome aforementioned issues through explicit segmentation of the activities of interest.
The recent adoption of recurrent (deep) neural networks for HAR applications has also led impressive recognition results, but
these methods come with their own set of problems \cite{hochreiter2001gradient}. 
For example, Long Short Term Memory (LSTM \cite{hochreiter1997long} models can learn infinite temporal contexts.
However, it is not reasonable to assume that an event in the distant past would actually influence current events.
Vanilla modeling is not able to capture such aspects.
This is even more pertinent for HAR problems as there is typically only little, if any, relation between current and distant past activities \cite{Bulling:2014:THA:2578702.2499621}.

Such observations lead us to the question about what would be the temporal context that is actually relevant for a model to consider in order to successfully represent activities of interest, and whether a model could make such a decision automatically.
If that was the case, then an externalization of such a data-driven decision regarding the relevant temporal context would lead to  insights about the analyzed data, possibly up to improved segmentation procedures.
Ultimately, we aim for a model to automatically learn its relevant temporal context.


In this paper we employ attention models to human activity recognition problems.
Essentially, attention models help a model learn a set of weights over a set of representations--data input--which signify the relative importance of each of the representations. 
For the case of activity analysis, these models would learn the contributions of all previous sensor readings that are considered for the analysis of a sample.
We use attention models for supervised learning tasks in HAR giving the model the ability to generate weight distributions over the history of a sample.
In doing so the model is incentivized to generate weights which place the weight on the context that is relevant for a classification decision. 

We explore the potential of attention models by adding an attention layer to a state-of-the-art, deep learning based HAR model \cite{ordonez2016deep}. 
We evaluate our approach on standard benchmarks (Opportunity, PAMAP2, Skoda), and 
results demonstrate significantly increased performance over current approaches, which emphasizes the relevance of the proposed approach.
We further explore what the models have learned by visualizing the attention model's weights, which provides additional insights into the model behavior. 



\section{Background}
Recent work in sequence modeling in HAR has mainly focused on convolutional networks (CNN), and on recurrent models such as LSTMs.
The attraction of CNNs lies in the fact that end-to-end learning is possible due to their stacked filtering layers that automatically capture hierarchical feature representations of the data.
Combined with clever combinations of pooling, that is, subsampling, and linear layers, very powerful recognition systems have been realized \cite{bhattacharya2016sparsification,zeng2014convolutional,yang2015deep}.
CNNs are applicable for analyzing sequential data only because of the sliding window trick, that slices out--typically fixed size--analysis frames from the input sequence of sensor data and promotes these through the network independently.
Some recent work has randomized this sliding window procedure in order to generate data variability that is exploited in Ensemble based approaches \cite{Guan:2017:EDL:3120957.3090076}.
However, the general sliding window principle remains unchanged.

Recurrent Models have also been applied very successfully in challenging HAR scenarios.
The vast majority of these approaches is based on--variants of--LSTMs \cite{hochreiter1997long}.
Such models incorporate specific gates into individual cells that allow for keeping an internal memory by feeding back a cell's output and by keeping track of the internal state.
\cite{hammerla2016deep} extensively analyzed the behavior of deep learning models in the wider HAR context, and one of the most promising current models represents a combination of CNNs--for representation learning--and LSTMs--for sequence learning: DeepConvLSTM \cite{ordonez2016deep}.

\vspace{-0.25em}
\section{Attention for HAR}
Previous deep learning approaches have focused on representing and modeling a fixed size temporal context for all sensor readings. 
Arguably, this approach works very well as such models currently dominate the most challenging HAR benchmarks (such as Opportunity \cite{chavarriaga2013opportunity}).
However, especially such challenging tasks exhibit both a substantial intra- as well as inter-class variance with regards to durations of the activities (cf.\ \cite{Guan:2017:EDL:3120957.3090076} for a detailed analysis of current benchmark datasets).
As such, using fixed window lengths will not naturally lead to ideal modeling and hence classification performance.

\begin{figure}[t]
	\centering
    \vspace*{-1em}
	\includegraphics[width=0.85\columnwidth]{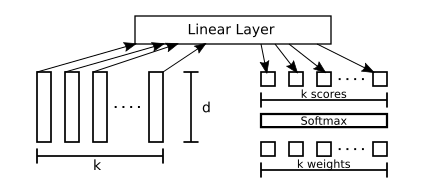}
    \vspace*{-1em}
    \caption{Illustration of adding attention (see text for description).} 
    \label{fig:attention_model}
    \vspace*{-1em}
\end{figure}

Instead, we explore how attention models can be employed for automatically determining the temporal context that is relevant for modeling activities.
In essence, such an approach would adapt the analysis windows in a data-driven manner.
Attention models have been introduced for natural language processing tasks for part of speech (POS) tagging \cite{kumar2016ask}.
The formal idea is that a set of linear layers and non-linearities are used to learn weights over $k$ vectors each of dimension $d$. 
Most architectures have a set of linear layers which map the dimension of these $d$-dimensional vectors to a one-dimensional score and these scores are then passed through the Softmax function to give the set of $k$ weights. 
The way each of these $k$ vectors is mapped to a one dimensional score is specific to the architecture. For instance, a linear layer could directly map the $d$-dimensional vector to a single dimension or one could add an intermediate linear layer to initially map the $d$-dimensional vector to, for example, a dimension $d/2$ and then a subsequent linear layer to map the $d/2$ dimensional vector to the one dimensional score. 
Fig.\ \ref{fig:attention_model} illustrates the general principle of adding attention to a deep learning model.

\begin{figure*}[t]
\centering
  \includegraphics[width=1.8\columnwidth]{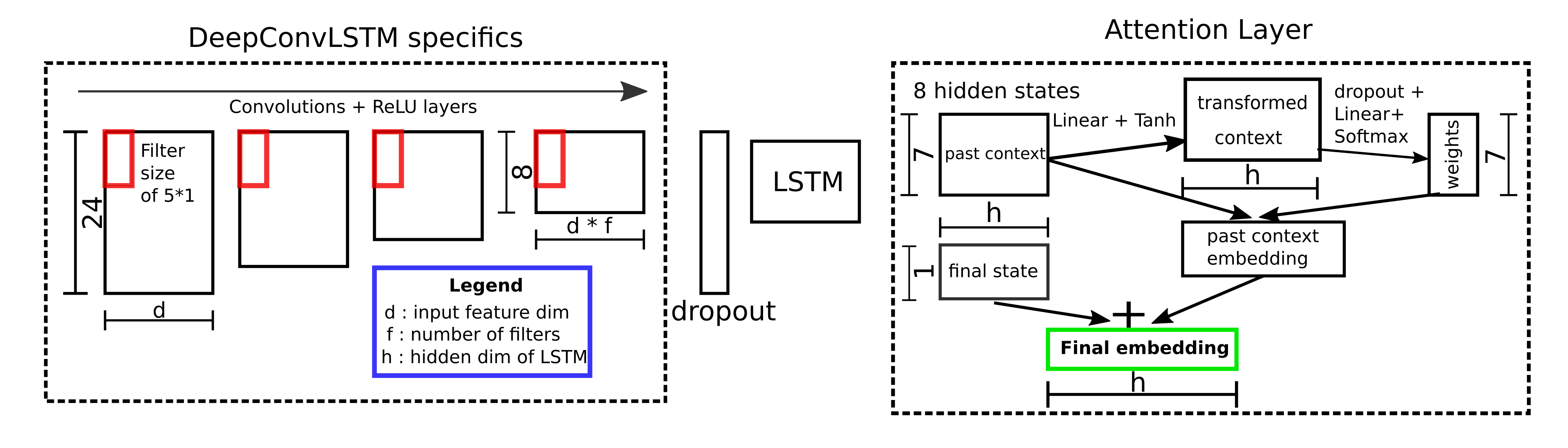}
  \caption{Adding attention to human activity recognition based on the DeepConvLSTM Architecture. The final embedding, highlighted in green, is used for prediction as opposed to the final hidden state in the original model (see text for description).}
  
%
\label{fig:overview}
\end{figure*}

We construct our HAR models by adding an attention layer to the state of the art architecture from \cite{ordonez2016deep} -- DeepConvLSTM (see below for model details). 
The general idea is to start with a large enough temporal context (sliding window) that was used in previous work and led to reasonable recognition performance.
We then add an attention layer to automatically rescale the weights of all samples in the analysis frame according to their relevance for modeling, which is according to our hypothesis that not all historic samples in an analysis frame are of (the same) relevance for modeling.
This relevance weighting of a sensor reading's history is a direct outcome of the attention layers, which we exploit for improving HAR models.
We do not change any other (hyper-) parameters 
to focus our exploration on the effect that introducing attention has on state-of-the-art activity recognition.
Fig.\ \ref{fig:overview} illustrates this architecture. 
Details are given in what follows.
%
%

\subsection{DeepConvLSTM and Attention }
DeepConvLSTM models \cite{ordonez2016deep} represent the state-of-the-art for deep learning based HAR applications, which motivates us to explore the addition of attention layers to this model architecture.
In this architecture the input, which is a window containing one second of data ($24$ samples for Opportunity \cite{chavarriaga2013opportunity}), is fed into four consecutive convolution layers with ReLU non-linearity. 
Through employing the windowing procedure, the sensor data input (time-series) is converted into a two-dimensional representation where the first dimension represents time and the other captures features -- $24 \times d$, where $d$ denotes the number of features in each sample.
Convolution filters are one-dimensional along the time axis and after the four convolution layers and successive downsampling through pooling the data representation is expanded to a two-dimensional arrary of size $8 \times d \times f$, where $d$ is the number of features and $f$ denotes the number of filters.
The latter was set to $64$ for each layer \cite{ordonez2016deep}.
This sequence of $8$ resulting feature vectors is then modeled by a two-layer LSTM with $128$ hidden units. 
The final hidden layer of the LSTM represents the embedding of the input frame, which is fed into a linear layer and a softmax to produce the prediction for an input frame.  

To incorporate the attention mechanism, we analyze the $8$ hidden states from the LSTM that represent the embeddings for the different parts of an input frame. 
We then consider the first $7$ hidden states as the historical temporal context and learn $7$ weights corresponding to these hidden states:
%

{\small
\begin{eqnarray}
\text{past context} & = & [h_1,h_2,h_3,...h_7] \label{eq:past_context} \\
\quad \text{current}  & = & h_{8} \label{eq:current_context} \\
\text{transformed context} & = & \tanh(W_1 \times \text{past context} + b_1) \label{eq:attention} \\
\text{weights} & = & \text{softmax}(W_2 \times \text{transformed context} \label{eq:weights}
 + b_2)  \\
\text{final embedding} & = & \text{past context} \times \text{weights} + \text{current} \label{eq:final_embeddings}
\end{eqnarray}
}

$b_1$ and $b_2$ denote the biases in the two linear layers, and $W_1$ and $W_2$ represent the 2D matrices in the linear layers.
We initially apply a linear transformation accompanied by a $\tanh$ linearity transforming each of these seven vectors of size $128$ into seven new vectors of size $128$ (Eq.\ \ref{eq:attention}). 
Another linear transformation converts these $7$ vectors of size $128$ into $7$ vectors of size $1$ essentially giving us scores for each of the hidden states. 
These scores are then passed through a softmax to give the final set of weights (Eq.\ \ref{eq:weights}). 
These weights are used to calculate a weighted sum of all the $7$ hidden states to give the final embedding for the past context. 
This past context is added to the last hidden state to give the final embedding for the input frame (Eq.\ \ref{eq:final_embeddings}). 
This final embedding is used for classification as opposed to the last hidden state used by DeepConvLSTM.

Note that the addition of the last hidden state to the embedding of the past context can be interpreted as a skip connection from the recurrent layers to the attention layer. 
Considering the computational graph that corresponds to this model, we observe that the model may decide to propagate the gradient only to the recurrent layers and could avoid the attention layers altogether. 
This is actually beneficial for HAR as datasets are often relatively small overfitting needs to be avoided, which could be realized explicitly through aggressive regularization, through the dropout layers \cite{srivastava2014dropout} shown in Fig.\ \ref{fig:overview} , or implicitly through these skip connections.

\section{Experiments}
Our explorations of the benefits that attention models may bring to human activity recognition are based on experimental evaluations on standard datasets from the field: Opportunity \cite{chavarriaga2013opportunity}, PAMAP2 \cite{reiss2012introducing}, and Skoda \cite{stiefmeier2008wearable}.
These datasets are very diverse in terms of the nature of activities and the relative distribution of activities. 
Therefore, the datasets present robust benchmarks for evaluating HAR systems. 
We employ standard training and evaluation protocols based on hold out datasets as they have been defined in the original publications (and summarized, e.g., in \cite{Guan:2017:EDL:3120957.3090076}).


All models were trained using the PyTorch deep learning framework \cite{paszke2017pytorch}. 
For all experiments a sliding window procedure was used to extract the processing frames our analysis is based on. 
Initial frame lengths were set to one second of data each, with an overlap of $50\%$ between consecutive frames.
Extracted frames were randomly shuffled during training to avoid bias.
All studied models produce sample-wise predictions, which is--in contrast to frame-wise prediction--more realistic for practical applications \cite{hammerla2016deep,Guan:2017:EDL:3120957.3090076}.

Models were trained using cross-entropy loss. 
Learning rate was fixed at $0.001$ and decayed after every epoch. Learning decay rate and the dropout values were optimized for all models, which seemingly have substantial impact on recognition performance. 
RMSProp was employed for optimization \cite{tieleman2012lecture}. 
Batch size was set to $100$ for all experiments and dropout layers were used for regularization. 
All code along with the best model weights for each of the datasets and the best hyper-parameters is available on our github page for reference\footnote{For reviewing purposes an anonymized archive of our github can be found at: \url{tinyurl.com/ybs4ndlv}}.
%
%
%

\vspace{-0.25em}
\section{Results}
Given the imbalance in class-distributions for all three datasets, we report results as mean f1 scores.
Statistical significance tests are based on Wilson score interval with 95\% confidence.
Recognition results for all benchmark datasets are given in Table \ref{tab:table1}.
It can be seen that the incorporation of attention models leads to significant increase in performance over the state of the art for both Opportunity and PAMAP2.
For Skoda we only see marginal improvements when introducing attention, which is similar to what has been reported in the literature for (other) model evaluations on this datasets.
As such, it may be concluded that by now a performance level has been reached for Skoda that bears no potential for further improvements.


\begin{table}[tb]
  \centering
    \caption{Sample-wise recognition results. Significant improvements over non-attention baselines are highlighted in bold (Wilson).}
    \vspace*{0.5em}
  \begin{tabular}{l r r r}
    Modeling  & \multicolumn{3}{c}{Datasets} \\
    Variant & Opportunity & PAMAP2 & Skoda \\
    \midrule
    DeepConvLSTM \cite{ordonez2016deep} & 67.2 & 74.8 & 91.2 \\
    b-LSTM-S \cite{hammerla2016deep} & 68.4 & 83.8 & 92.1 \\
    \midrule
    \textbf{Att. Model} & \textbf{70.7}  & \textbf{87.5}  & 91.3  \\
    Confidence Interval & $\pm$ \textbf{.003} & $\pm$ \textbf{.002} & $\pm$ .004
%
%
%
%
  \end{tabular}
  \label{tab:table1}
  \vspace*{-1em}
\end{table}

\section{Discussion}
\begin{figure}[tb]
\centering
	\vspace*{-1em}
  \includegraphics[width=0.9\columnwidth]{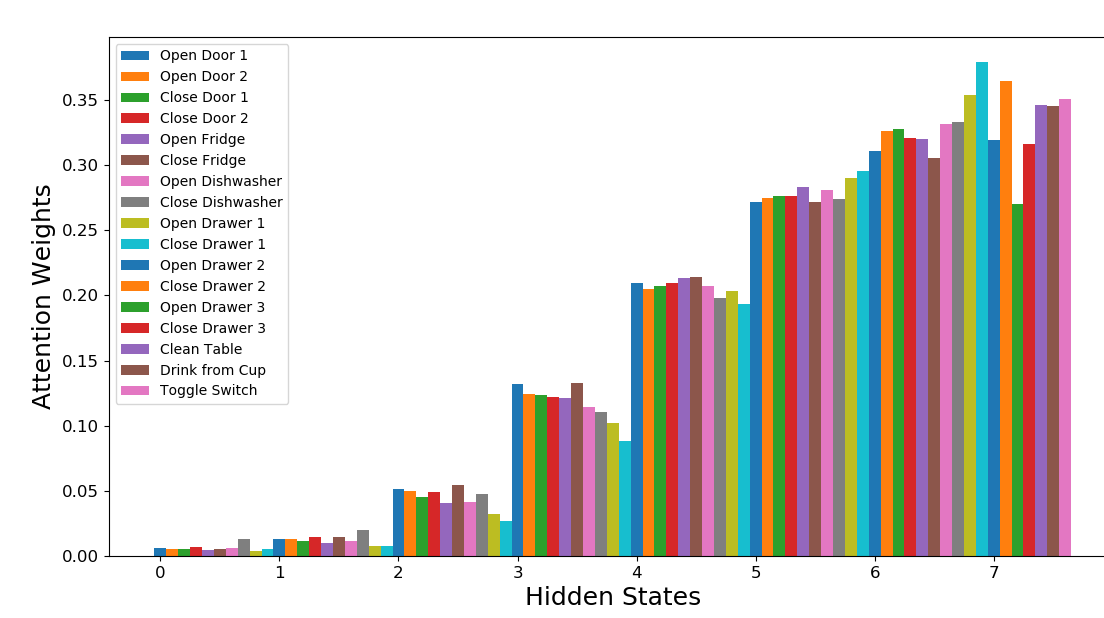}
  \caption{Visualizing weights learned by the best attention model on the Opportunity test dataset (best viewed in color).}
  \vspace*{-2.5em}
~\label{fig:attn_vis}
\end{figure}

Figure \ref{fig:attn_vis} visualizes the weights of the best model. While evaluating, each sample has a set of 7 weights associated with the relative importance of the first 7 hidden states of the LSTM. We take the median of these weights across all samples belonging to a certain activity. 
The visualization shows interesting insights into the model's behavior. We notice that most of the weight is concentrated on the last few hidden states. This is reasonable as 
these states capture the summary of the input frame and through the LSTM recurrence the last hidden states capture more information than  previous ones, and hence have a more dominant contribution to the final context embedding. 
However, only using the final hidden state--as LSTMs do--is detrimental as there may be some important information at the start of the input frame. 
Therefore, through using attention models we see a significant amount of the weight being placed on the past hidden states as well and this allows the model to capture the context more effectively as opposed to only relying on the last hidden state of the LSTM. 
The improvements in recognition performance confirm the benefit of adding attention to deep, recurrent HAR models.

We also observe that for all activities analyzed, the weight on the first two hidden states is close to zero.
This is likely due to the first few hidden states not yet being able to capture anything valuable because the history at this stage is too short and thus rather uninformative.
The attention model explicitly downweights those initial states as they do not contribute much to the model and thus rather "waste" model parameters if included.
In summary, the attention mechanism effectively shrinks and focuses the history of a sensor reading that a HAR model needs to focus on.

We also observe that among all the (Opportunity) activities, the activity "Open Door 3" has the most spread out weights on all the hidden states. This is interesting because it suggests that this activity might involve multiple distinct segments as the model distributes the weight evenly on hidden states. On further inspection, we realize that this activity is about opening the lowest drawer in a cupboard containing three drawers and hence one might need to perform multiple smaller activities such as bending down, opening the drawer and rising up to perform this activity. Therefore, the model is incentivized to distribute the weight more evenly to capture these sub-activities.     
This last aspect is the basis for future developments and applications as--essentially--it is the starting point for novel segmentation schemes.

\bibliographystyle{SIGCHI-Reference-Format}
\bibliography{sample}

\end{document}